\documentclass[lettersize,journal]{IEEEtran}
\usepackage{amsmath,amsfonts}
\usepackage{algorithmic}
\usepackage{algorithm}
\usepackage{array}
\usepackage[caption=false,font=normalsize,labelfont=sf,textfont=sf]{subfig}
\usepackage{textcomp}
\usepackage{stfloats}
\usepackage{url}
\usepackage{verbatim}
\usepackage{graphicx}
\usepackage{cite}
\usepackage{color}
\usepackage{hyperref}
\hypersetup{colorlinks=true, linkcolor=blue}
\usepackage{times}
\usepackage{epsfig}
\usepackage{graphicx}
\usepackage{amsmath}
\usepackage{amssymb}
\usepackage{booktabs}
\usepackage{multirow}

\begin{document}

\title { 
Coding-Prior Guided Diffusion Network for Video Deblurring
}

\author{
        ~Yike~Liu,
        ~Jianhui~Zhang,
        ~Haipeng~Li,
        ~Shuaicheng~Liu,
        and Bing~Zeng}

\maketitle

\begin{figure*}[!t]
  \centering
  \includegraphics[width=0.95\textwidth]{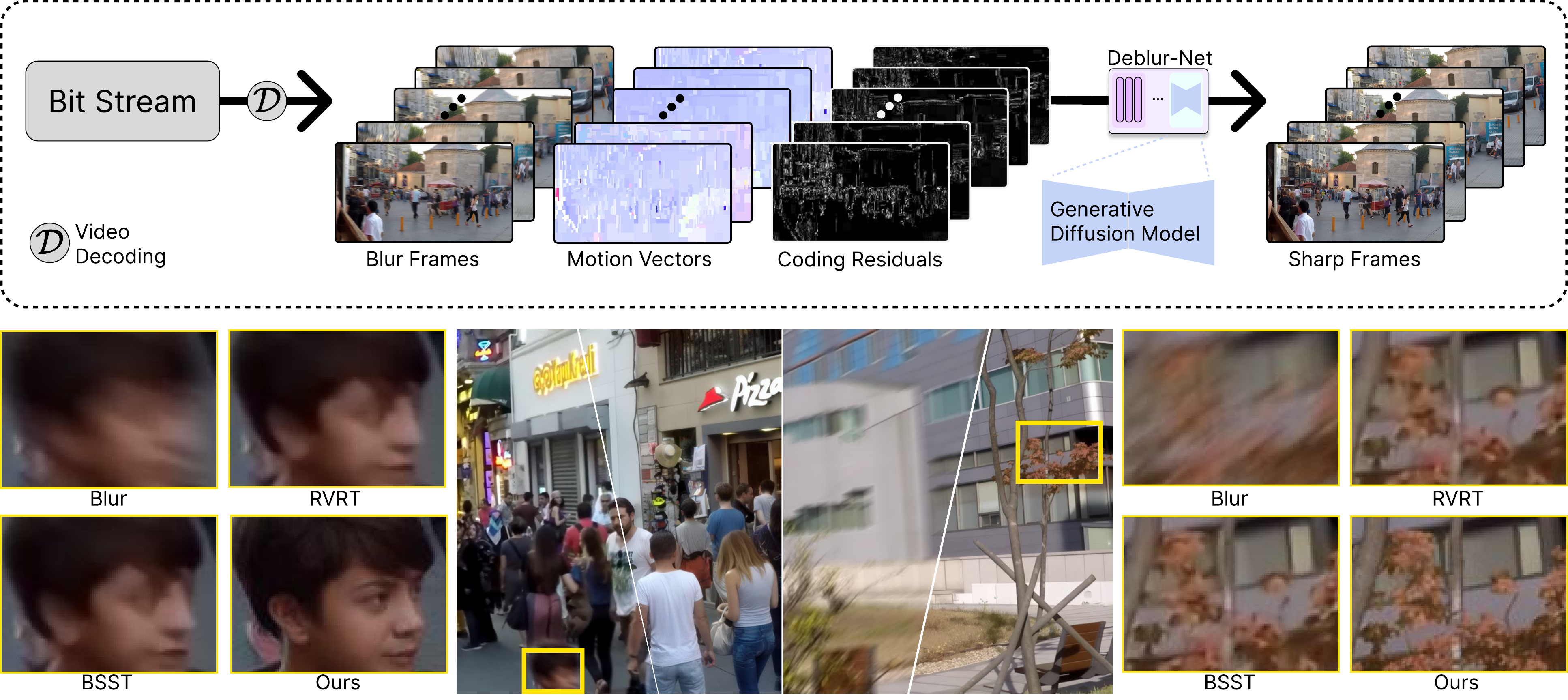}
  \caption{
  Our framework integrates video decoding priors with diffusion-based generative priors for video deblurring. From the compressed video bit stream, we extract motion vectors (MVs) and coding residuals (CRs) alongside decoded frames. Comparative results show improved blur removal and detail reconstruction. (\textbf{Zoom in for best view.})}
  \label{fig_teaser}
\end{figure*}

\begin{abstract}

While recent video deblurring methods have advanced significantly, they often overlook two valuable  prior information: (1) motion vectors (MVs) and coding residuals (CRs) from video codecs, which provide efficient inter-frame alignment cues, and (2) the rich real-world knowledge embedded in pre-trained diffusion generative models. We present CPGD-Net, a novel two-stage framework that effectively leverages both coding priors and generative diffusion priors for high-quality deblurring. First, our coding-prior feature propagation (CPFP) module utilizes MVs for efficient frame alignment and CRs to generate attention masks, addressing motion inaccuracies and texture variations. Second, a coding-prior controlled generation (CPC) module network integrates coding priors into a pre-trained diffusion model, guiding it to enhance critical regions and synthesize realistic details. Experiments demonstrate our method achieves state-of-the-art perceptual quality with up to 30\% improvement in IQA metrics. Both the code and the coding-prior-augmented dataset will be open-sourced.

\end{abstract}


\section{Introduction}

Video deblurring aims to recover sharp images from videos blurred by camera or object motion. This technology is highly valuable as it not only improves visual quality but also benefits many downstream applications like tracking~\cite{tracking}, video stabilization~\cite{video_stabilization_1,video_stabilization_2}. Real-world videos typically require compression due to their large data size. Modern video codecs~\cite{H.264,HEVC,VVC} effectively reduce redundancy by utilizing inter-frame correlations, generating useful intermediate data including motion vectors and residual coefficients. While these encoding features effectively represent inter-frame relationships, they have been overlooked in previous deblurring research. Similarly, the rich real-world knowledge in current pre-trained generative models hasn't been utilized for video deblurring. Our work improves video deblurring by effectively using both types of prior information.

Previous work in video deblurring has made significant progress through deep learning, particularly by exploiting inter-frame information. Flow-guided methods~\cite{pvdnet,stdanet,basicvsr++,memdeblur,rvrt} achieve promising results by using bidirectional optical flow to direct deformable convolution operations for feature alignment. However, those approach necessitate additional computational overhead for optical flow estimation. Meanwhile, transformer-based architectures~\cite{vrt,rvrt,bsst} demonstrate strong performance through self-attention mechanisms, but their performance remains constrained by the intrinsic information available in the input sequences, limiting perceptual quality.

The concept of inter-frame correlation is fundamental to video coding, where motion compensation prediction~\cite{mcp1,mcp2} generates two complementary signals: motion vectors (MVs) that encode block displacements, and coding residuals (CRs) that preserve uncompensated details. Notably, unlike computationally intensive optical flow, MVs offer "free" motion cues extracted directly from compressed streams. Prior work has demonstrated their effectiveness in temporal correspondence tasks, from homography estimation~\cite{codinghomo} to optical flow computation~\cite{mvflow}. For video deblurring specifically, MVs provide reliable alignment cues, while CRs complement this process by preserving unpredicted details and texture variations. Crucially, CRs inherently identify challenging regions such as motion-induced blur and sharp-to-blur transitions—precisely where traditional alignment algorithms struggle. By jointly leveraging these coding priors, our approach achieves both efficient frame matching and targeted attention to critical variation regions.

While motion compensation-based approaches can effectively reduce blurring artifacts, their output quality remains inherently limited by the input video's characteristics. Recent advances in denoising diffusion probabilistic models (DDPMs)~\cite{ddpm} have shown remarkable capabilities in text-to-image~\cite{stablediffusion,controlnet} and text-to-video generation~\cite{sora}. However, existing diffusion-based video deblurring methods~\cite{vddiff} primarily use diffusion models as intermediate priors, failing to fully exploit their native image generation capacity. To overcome this limitation, we integrate a pre-trained diffusion foundation model~\cite{stablediffusion}, directly leveraging its learned priors of high-quality real-world images for enhanced video deblurring.

In this work,we propose a novel two-stage framework, CPGD-Net, which effectively utilizes both coding priors and generative diffusion priors for video deblurring. Departing from conventional approaches that process only decoded frames, our method additionally extracts and utilizes encoding priors—specifically motion vectors and residual coefficients—which are jointly processed with the video frames, serve as input to our network as shown in the first row of Figure~\ref{fig_teaser}. Technically, in the first stage, we design an coding-prior feature propagation(CPFP) module that performs initial alignment using motion vectors and generates high-quality attention masks from residuals to identify and compensate for motion vector inaccuracies, results in efficiently restores blurred videos by leveraging temporal information. The second stage incorporates coding-prior controlled generation (CPC) module to adding generative details. By integrating coding priors into the attention mechanism of the control network, the model can better focus on motion-affected regions and areas with texture variations, thereby producing higher-quality results. Through this two-stage architecture, we achieve optimal utilization of both coding priors and generative priors, yielding significantly improved perceptual quality as shown in the second row Figure~\ref{fig_teaser}. Overall, our primary contributions include:

\begin{itemize}
    \item To the best of our knowledge, we are the first to propose and implement a framework that systematically incorporates video codec priors for video deblurring task.

    \item We pioneer the application of text-to-image diffusion model priors in video deblurring to generate high-quality output frames.

    \item We design two novel components: (a) a coding-prior feature propagation module for efficient utilization of video codec information, and (b) a coding-prior control network to regulate the generative model's output.

    \item We augment standard benchmark datasets with corresponding video coding information and demonstrate superior perceptual quality in our experimental results.
\end{itemize}

\begin{figure*}[h]
  \centering
  \includegraphics[width=0.95\linewidth]{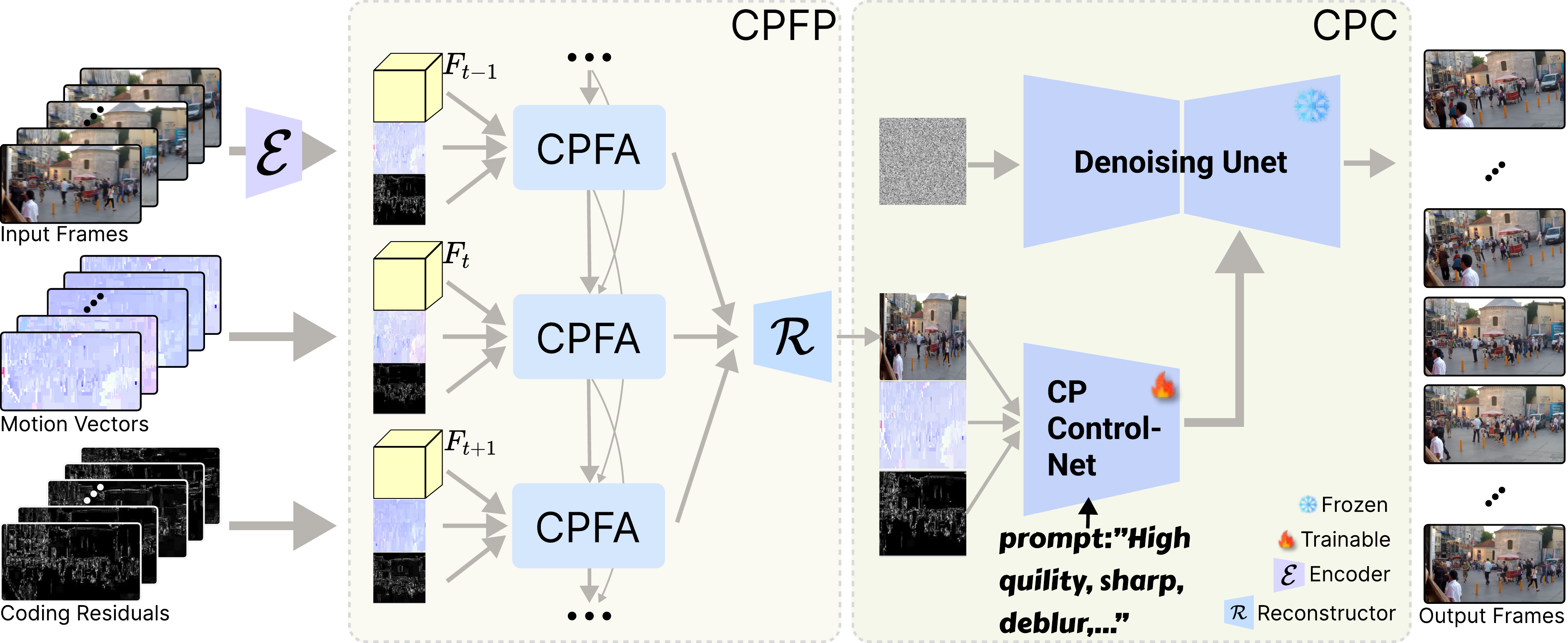}
  \caption{Overview of our CPGD-Net pipeline: Our framework consists of (1) a Coding-Prior Feature Propagation (CPFP) module that aligns features through cascaded CPFA blocks and performs initial restoration, followed by (2) a Coding-Prior Controlled Generation (CPC) module that synthesizes high-quality outputs by conditioning a diffusion model on the stage-one results, coding priors (motion vectors/residuals), and text prompts through a denoising process. }
  \label{fig_pipe}
\end{figure*}

\section{Relative Work}
\subsection{Video Deblurring}

Video deblurring approaches primarily fall into two categories: RNN based and transformer based methods. Early RNN based works like STRCNN~\cite{strcnn} and IFRNN~\cite{ifrnn} employed recurrent architectures for temporal modeling, while later advancements like STDANet ~\cite{stdanet} introduced flow-guided deformable convolution for alignment. Transformer based methods such as VRT~\cite{vrt} and RVRT~\cite{rvrt} leveraged spatiotemporal self-attention mechanisms, while BSSTNet~\cite{bsst} further improving efficiency through optical-flow-derived blur maps for sparse attention. 
Although VD-Diff~\cite{vddiff} successfully combined wavelet-aware transformers with diffusion models and achieve great performance, its application of the diffusion model was restricted to intermediate feature generation. While these methods have advanced the field, their performance remains constrained by input video quality, and none exploit coding priors or the generative capabilities of pre-train diffusion models 

\subsection{Diffusion Models}
Diffusion models have demonstrated remarkable success across visual computing tasks. DDPM~\cite{ddpm}  establishes denoising diffusion probabilistic models, followed by DDIM~\cite{ddim} that achieved breakthrough efficiency through non-Markovian inference, enabling 10-50 times speedup via deterministic sampling. Stable Diffusion~\cite{stablediffusion} implements latent-space compression to reduce computation by 5-10 times, and ControlNet~\cite{controlnet} enables conditional generation through parallel weight cloning. Significant progress has also been made in restoration-specific applications: DDNM~\cite{ddnm} develops null-space projection for blind restoration, Diff-IR~\cite{diffir} creates a unified inverse-diffusion framework, DiffBIR~\cite{diffbir} introduces a decoupled two-stage strategy combining degradation-aware denoising with detail enhancement, and Ren et al.~\cite{msgd_image_deblur} propose wavelet-domain multiscale diffusion with structural guidance. While these methods have shown excellent performance in image generation and restoration, their application to video deblurring, particularly when combined with coding priors, remains largely unexplored.

\subsection{Video Coding Prior}

Video coding priors, particularly motion vectors (MVs), demonstrate significant utility across various computer vision tasks. Earlier works successfully leverage these priors in multiple domains. For video stabilization, CodingFlow~\cite{codingflow} establishes an MV-based framework that bypasses traditional optical flow's computational overhead. In super-resolution, Chen et al.~\cite{compressed_super-resolution} enhance reconstruction by synergistically integrating decoding priors with deep learning. Zhang et al.~\cite{mv_action_recognition} implement real-time action recognition using MVs as an efficient optical flow alternative, while CPGA~\cite{cpga} employs coding priors for video quality enhancement. Tan et al.~\cite{mv_segmentation} develop compressed-domain propagation for accelerated segmentation, and MV-Flow~\cite{mvflow} improves optical flow estimation using MV priors. CodingHomo~\cite{codinghomo} introduces an unsupervised homography estimation framework that utilizes MVs to achieve robust performance in complex scenarios. Notably, despite their proven effectiveness in these diverse tasks, coding priors remain unexplored for video deblurring. Our work pioneers the application of coding priors to video deblurring, achieving superior performance.

\section{Methods}

\subsection{Motivation and Overview}

In this work, we propose an effective two-stage pipeline (as illustrated in Figure~\ref{fig_pipe}) that leverages two valuable types of prior information: coding priors and generative priors, to achieve high-quality video deblurring.

\begin{figure}[t]
  \centering
  \includegraphics[width=0.95\linewidth]{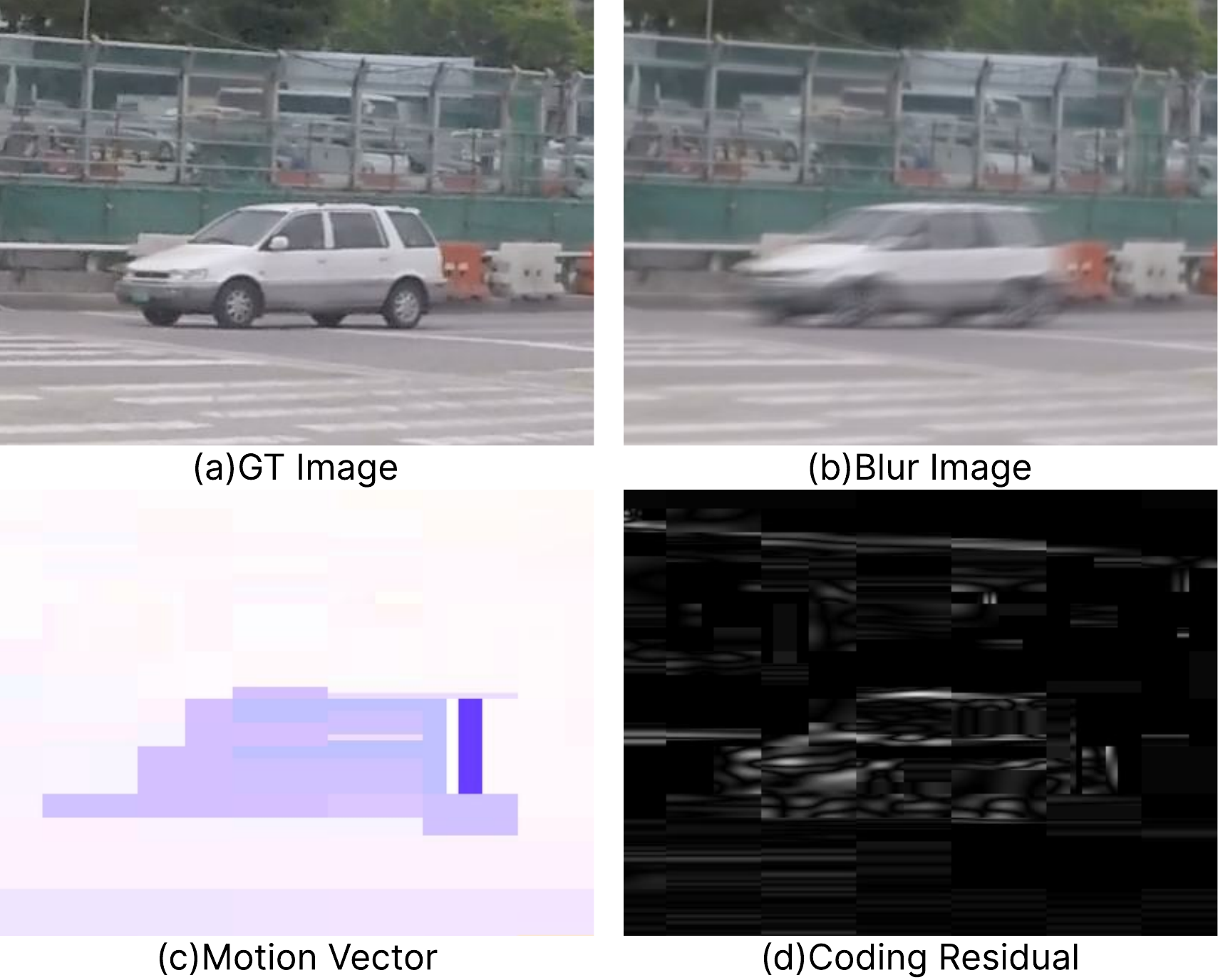}
  \caption{Visualization of MVs and CRs. (a) Ground truth image; (b) Motion-blurred image. (c) Motion vector. (d) Coding residual.}
  \label{fig_cp}
\end{figure}


Motion vectors explicitly capture inter-frame motion relationships, while coding residuals implicitly indicate the effectiveness of motion compensation. Regions with large coding residuals typically correspond to motion-induced blur, occlusions, or significant texture variations where block matching fails. For motion-blurred videos, these regions often represent either object motion blur or global camera motion blur. As illustrated in Figure~\ref{fig_cp},  we compare a sharp-blurred image pair with visualization of the motion vectors and residual signals associated with the blurred frame. These coding-derived priors demonstrate significant value for deblurring tasks, as they provide explicit motion representation and implicit compensation quality indicators. Consequently, we extract both the decoded frames and these intermediate coding variables (motion vectors and coding residuals) as prior information for subsequent processing.

Prior work~\cite{controlnet,diffbir} has demonstrated that generative networks possess remarkable capabilities for high-quality image synthesis. When combined with control networks, they show exceptional potential to enhance images by adding plausible and visually pleasing details while preserving original content. However, directly using blurred frames as control conditions makes it challenging for the network to extract meaningful features. To address this, we design a two-stage framework: Stage 1: We first encode video frames into a latent space, then employ a cascaded Coding-prior Feature Alignment (CPFA) blocks to efficiently align inter-frame information, followed by a feature fusion module to produce initially restored frames. Stage 2: We utilize both the Stage 1 outputs and coding priors as conditional inputs to a special designed coding-prior controlled generation (CPC) Module. This guides the generative model to reconstruct high-quality frames from noise while maintaining content fidelity with the original video. By combining these two stages, our framework achieves sota generative video deblurring results.

\subsection{Coding-Prior Feature Propagation Module}

\begin{figure}[t]
  \centering
  \includegraphics[width=\linewidth]{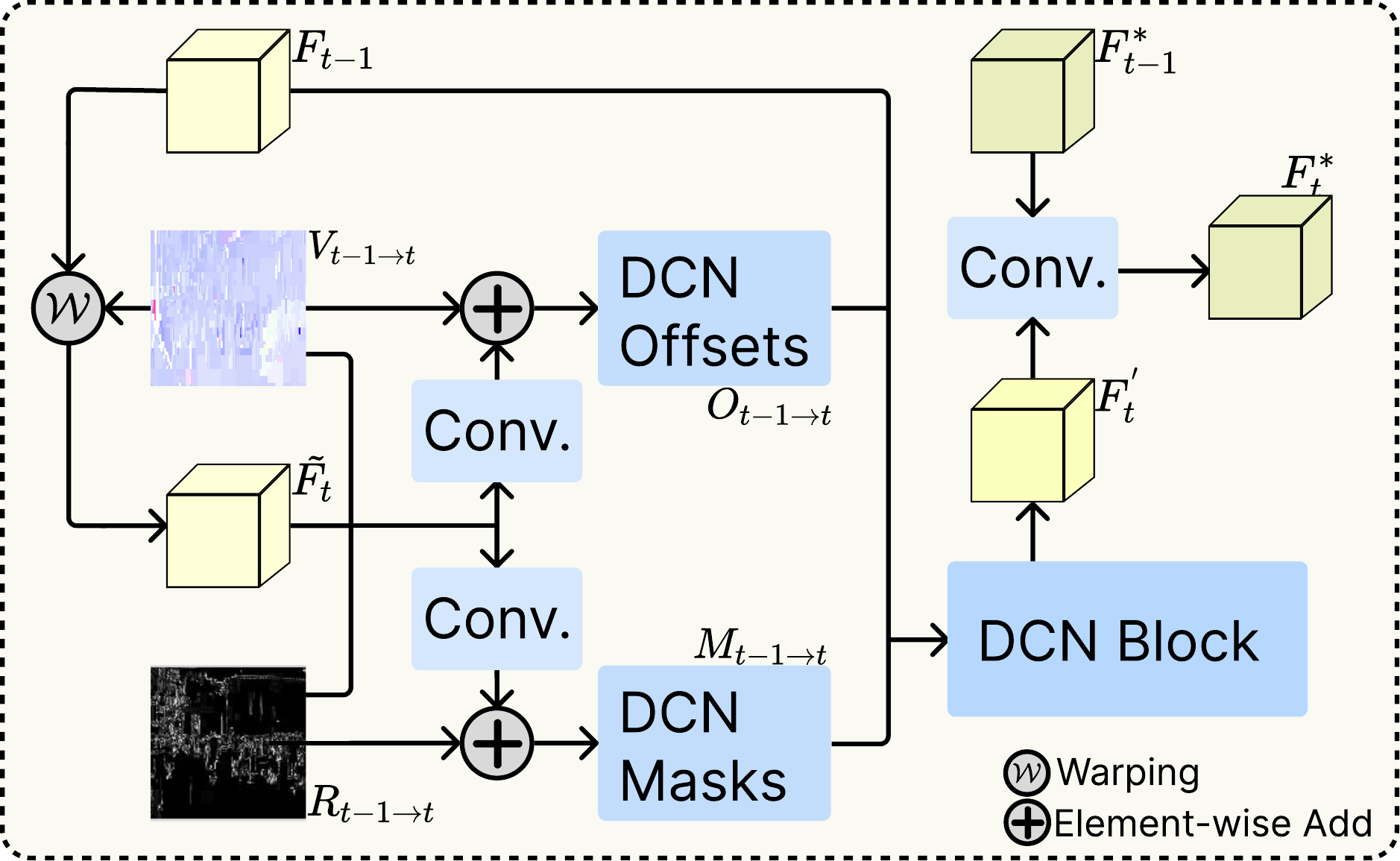}
  \caption{Detile of the CPFA block. Given features $F_{t-1}$, motion vectors $V_{t-1\rightarrow t}$, and coding residuals $R_{t-1\to t}$: (1) Warps $F_{t-1}$ using $V_{t-1\rightarrow t}$ to obtain $\widetilde{F}_t$;(2) Predict deformable convolution parameters $\{O_{t-1\rightarrow t}, M_{t-1\rightarrow t}\}$ via concatenated $\{V_{t-1\rightarrow t}, R_{t-1\to t}, \widetilde{F}_t\}$; (3) Apply $\mathrm{DCN}(F_{t-1}, O_{t-1\rightarrow t}, M_{t-1\to t})$ and fuse with $F^*_{t-1}$ to output $F^*_t$.
}
  \label{fig_cpfa}
\end{figure}

Inspired by prior work~\cite{basicvsr++}, we propose a Coding-Prior Feature Propagation (CPFP) module that leverages video coding priors to enhance video deblurring performance. The CPFP module consists of cascaded Coding-Prior Feature Alignment (CPFA) blocks, as illustrated in Figure~\ref{fig_cpfa}. 

Since motion vectors($V_{t-1\rightarrow t}$) inherently capture displacement relationships between frames, we employ them to warp previous frame features $F_{t-1}$ to the current temporal position $\widetilde {F_{t}}$ as:

\begin{equation}
\label{F_wrap}
\begin{split}
\widetilde {F_{t}} = \mathcal{W}(V_{t-1\rightarrow t}, F_{t-1}), 
\end{split}
\end{equation}
where $\mathcal{W}$ is warping operation.

We enhance alignment accuracy using deformable convolution ($\mathcal{D}$) with learned offsets $O_{t-1\rightarrow t}$ and an attention mask $M_{t-1\rightarrow t}$. Building upon the characteristics of coding priors, we specifically design the deformable convolution to focus on regions where basic alignment fails - precisely those areas exhibiting significant coding residual ($R_{t-1\rightarrow t}$) values. The coding residual $R_{t-1\rightarrow t}$ is normalized to the range [0,1] to ensure compatibility with neural network feature requirements. Offsets is generated by a convolution network ($\mathcal{C}_{o}$) which has input of concatenation of $V_{t-1\rightarrow t}$, $\widetilde {F_{t}}$ and $R_{t-1\rightarrow t}$ and then summing skip connected $V$ for the final $O_{t-1\rightarrow t}$ as:
\begin{equation}
\label{DCN_offset}
\begin{split}
O_{t-1\rightarrow t}&= V_{t-1\rightarrow t} + \mathcal{C}_{o}(V_{t-1\rightarrow t}, \widetilde {F_{t}}, R_{t-1\rightarrow t}), 
\end{split}
\end{equation}
and $M_{t-1\rightarrow t}$ is generated by adding skip connected $R_{t-1\rightarrow t}$ with the output of convolution network ($\mathcal{C}_{m}$) which also has input of concatenation of $V_{t-1\rightarrow t}$, $\widetilde {F_{t}}$ as:  
\begin{equation}
\label{DCN_mask}
\begin{split}
M_{t-1\rightarrow t}&= R_{t-1\rightarrow t} + \mathcal{C}_{m}(V_{t-1\rightarrow t}, \widetilde {F_{t}}, R_{t-1\rightarrow t}).
\end{split}
\end{equation}
The DCN is then applied to $F_{t-1}$ together with $O_{t-1\rightarrow t}$ and $M_{t-1\rightarrow t}$ as:

\begin{equation}
\label{DCN}
\begin{split}
F^{'}_{t} &= \mathcal{D}(F_{t-1}, O_{t-1\rightarrow t}, M_{t-1\rightarrow t}),
\end{split}
\end{equation}
where $\mathcal{D}$ denotes a deformable convolution.

The output $F^{'}_{t}$ combines with $F^{*}_{t-1}$ from previous CPFA block operations through an additional convolution network to produce the final temporal feature $F^{*}_{t}$.

Our module substitutes the commonly used optical flow with MVs, eliminating additional computational costs. Meanwhile, the incorporation of coding residual enables the deformable convolution to focus more effectively on blurred regions during both offset computation and final output generation, achieving superior feature alignment for blurred areas.

\subsection{Coding-Prior Controlled Generation Module}

\begin{figure}[t]
  \centering
  \includegraphics[width=\linewidth]{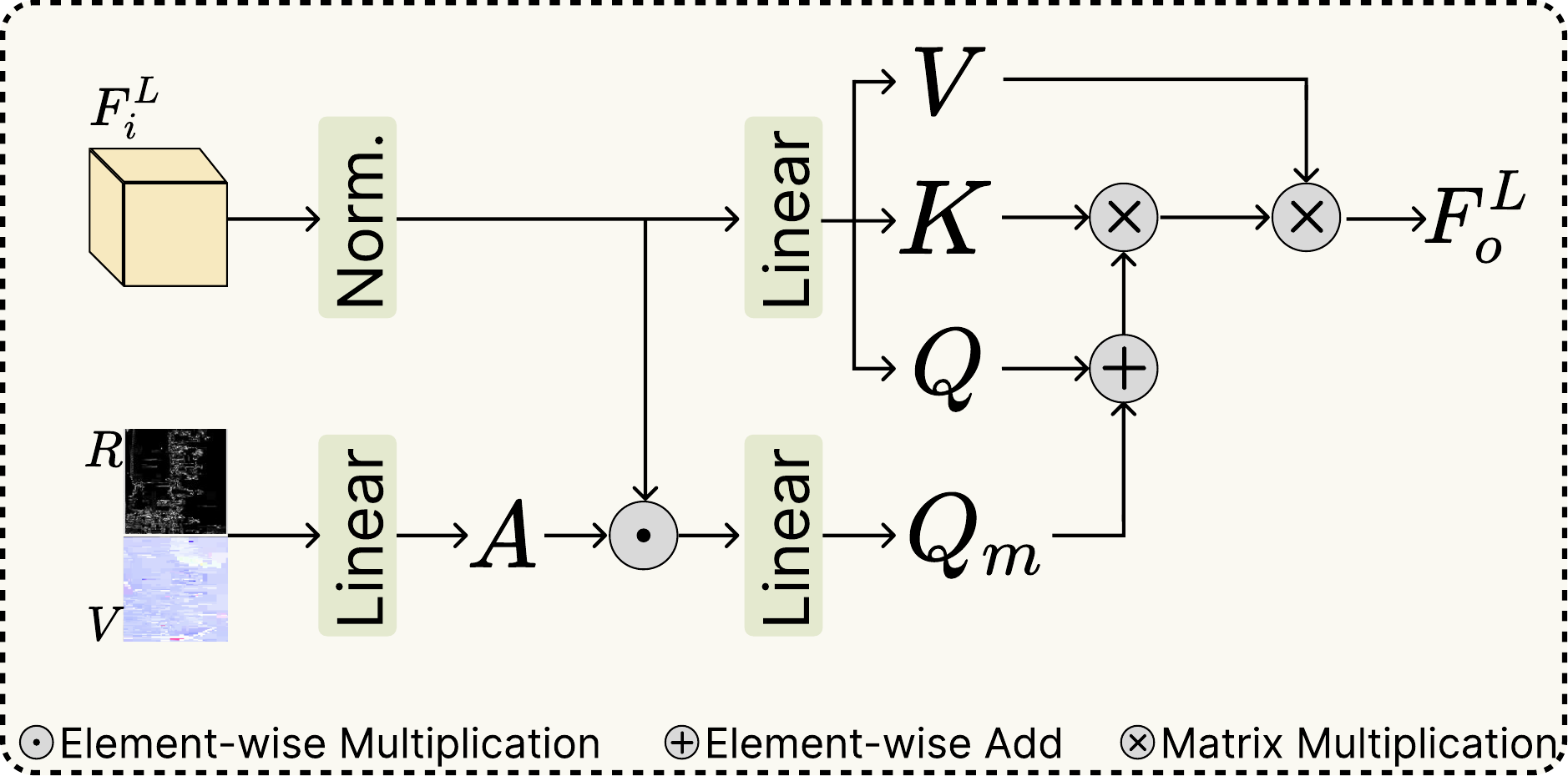}
  \caption{ At each transformer layer, motion vectors $V$ and coding residuals $R$ are converted to an attention mask $A_i$. The mask modulates the query $Q$ to prioritize blur-sensitive regions.
}
  \label{fig_cpc}
\end{figure}

While Stage-1 successfully restores sharp structural information, its outputs $X$ often lack high-frequency details due to irreversible information loss in motion-blurred inputs. To address this limitation, we propose a Coding-Prior Controlled Generation (CPC) module that leverages a pre-trained generative foundation model~\cite{stablediffusion}, enhanced with our novel CPControlNet. The CPControlNet effectively integrates coding priors (motion vectors $V$ and coding residuals $R$) with text prompts $P_t$ to guide the generation process from intermediate state $y^t$ to refined output $y^{t-1}$. This synergistic combination enables the generation of results with significantly enhanced visual fidelity while maintaining temporal consistency. The overall generation process is formally defined as:
\begin{equation}
\label{generation}
\mathcal{G}_t:(X,V,R,P{t},y^t) \rightarrow y^{t-1},
\end{equation}
where $\mathcal{G}$ denotes our enhanced generation framework.

Guided by our earlier findings that $V$ and $R$ provide critical priors for deblurring, we maintain their incorporation into the ControlNet's attention architecture as shown in Figure~\ref{fig_cpc}. This ensures focused attention on regions affected by motion or significant temporal texture variations. Firstly, at the i-th level of transformer, $V$ and $R$ are converting into mask $A$ through a linear layer ($\mathcal{L}_m$). Element-wise multiplying $A$ with the latent feature ($F_{i}^{L}$) as input for generating a mask-guided attention query ($Q^m$), which is additively combined with the standard self-attention query ($Q$) to yield a refined attention map ($Q^{'}$) as: 

\begin{equation}
\label{Q}
\begin{split}
A &= \mathcal{L}_m(V,R), \\
Q^{'} &= \mathcal{L}_q(F_{i}^{L}) + \mathcal{L}_{qm}({F_{i}^{L}} \odot {A}),
\end{split}
\end{equation}
where $\mathcal{L}_q$ is the query projection, and $\mathcal{L}_{qm}$ processes mask-modulated query. This directs stronger attention to motion-distorted regions. Finally, the output feature $F^L_o$ is formulated as:
\begin{equation}
\label{attention}
\begin{split}
K &= \mathcal{L}_k(F_{i}^{L}),\\
V &= \mathcal{L}_v(F_{i}^{L}),\\
F^L_o &= \text{Softmax}(\frac{Q_{i}^{'}K^T}{\sqrt{d}})V_i.
\end{split}
\end{equation}
where $\mathcal{L}_k$ is the key projection, and $\mathcal{L}_v$ is the value projection.

The output of CPControlNet is injected into the denoising U-Net. Using $X$, $V$, $R$, and $P_{t}$ as conditioning inputs, our framework produces semantically coherent, high-detail $y^0$ that faithfully preserve input semantics, exhibit superior temporal stability and recover photorealistic textures through generative refinement.

\section{Experiment}

\subsection{Datasets}

\textbf{GoPro Dataset}~\cite{gopro} contains 3,214 blurry-sharp image pairs at 1280×720 resolution, partitioned into 2,103 training pairs and 1,111 testing pairs.

\textbf{DVD Dataset}~\cite{dvd} comprises 71 dynamic videos yielding 6,708 blurry-sharp frame pairs, with 5,708 pairs from 61 videos allocated for training and 1,000 pairs from 10 videos designated for testing.

\textbf{HQ Video Dataset} The curated filtered high-quality video dataset comprises two standardized video coding test collections: CTC~\cite{ctc} and MCML~\cite{mcml}. It contains 18 raw video sequences totaling 8980 frames, with resolutions ranging from 4K (3840×2160) to 720p (1280×720) in YUV420 format. As the sequences remain uncompressed, the dataset preserves uncompromised image quality without coding artifacts.

\subsection{Implementation Details}

\begin{figure}[t]
  \centering
  \includegraphics[width=1\linewidth]{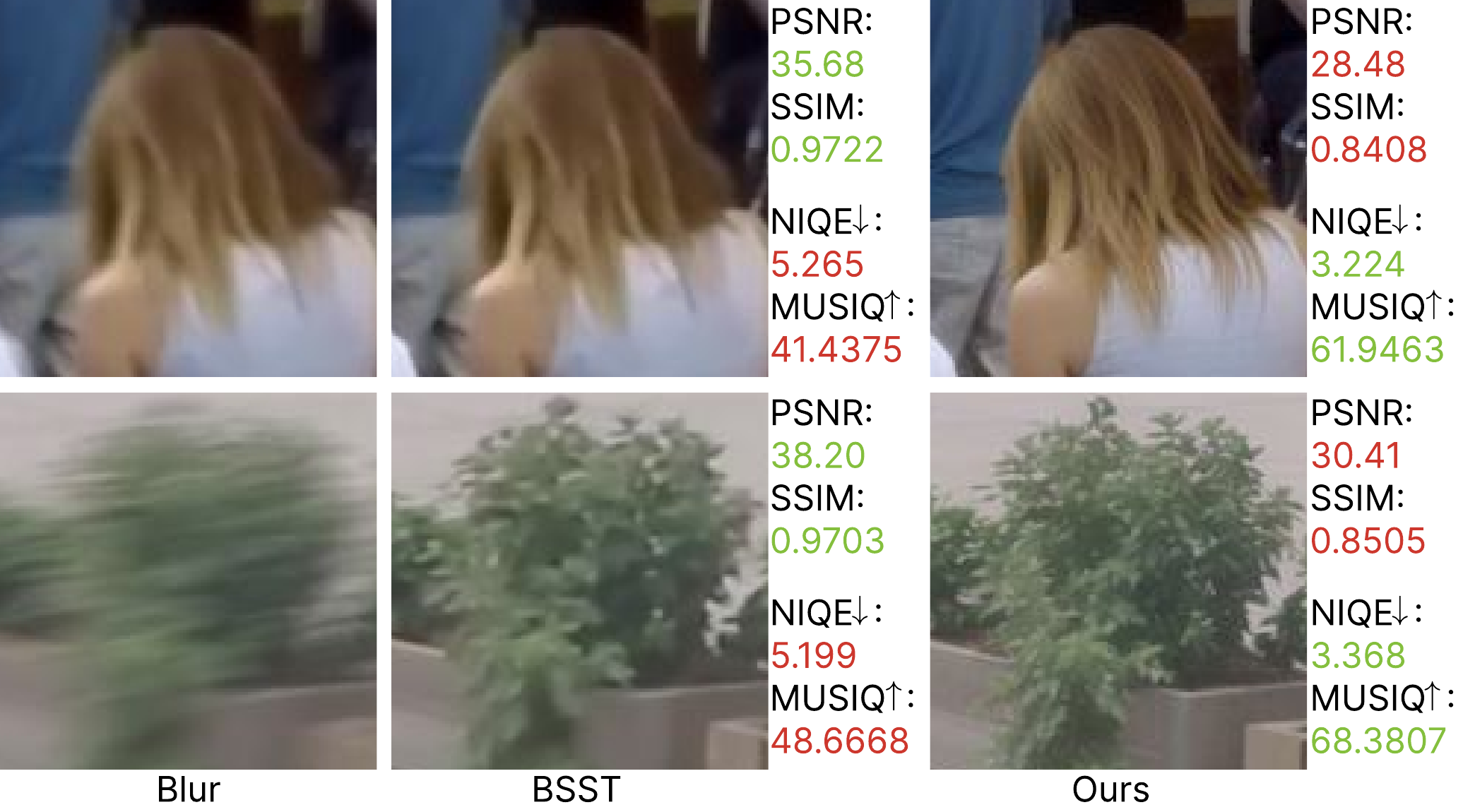}
  \caption{\textbf{Discrepancy between metrics and perceptual quality.} Our CPGD produces visually superior textures despite lower PSNR/SSIM scores, demonstrating the necessity of no-reference metrics (NIQE/MUSIQ) for quality assessment.}
  \label{fig_psnr}
\end{figure}

Our framework is implemented using PyTorch~\cite{pytorch}. We perform encoding operations on the dataset to extract corresponding coding priors through VVC~\cite{VVC}. The training procedure is divided into two distinct phases, executed across 8 NVIDIA A100 GPUs.

Phase I trains the CPFP module with a batch size of 8 and an initial learning rate of $1\times10^{-4}$. Input images are randomly cropped into 256×256 patches. This phase employs the Adam optimizer~\cite{adam} with $L1$ loss, configured with $\beta_1 = 0.9$ and $\beta_2 = 0.999$.

Phase II trains the CPC module while keeping both the denoising UNet and VAE components fixed (initialized from Stable Diffusion 2.0~\cite{stablediffusion}) to preserve their pretrained generative capabilities. The CPControlNet is fine-tuned from IRControlNet~\cite{diffbir}, with enhanced control generation through video information integration. To fully leverage the pretrained model's capacity, we use 512×512 crops during this phase, maintaining a batch size of 8 and the learning rate of $1\times10^{-5}$. For improved perceptual quality, we create a hybrid training dataset combining: Original deblurring samples and degraded samples from HQ Video Dataset processed through CodeFormer~\cite{codeformer}.To enable efficient inference, we implement a spaced DDPM sampling schedule~\cite{spacesample} requiring only 50 sampling steps.

\begin{figure*}[!ht]
  \centering
  \includegraphics[width=1\linewidth]{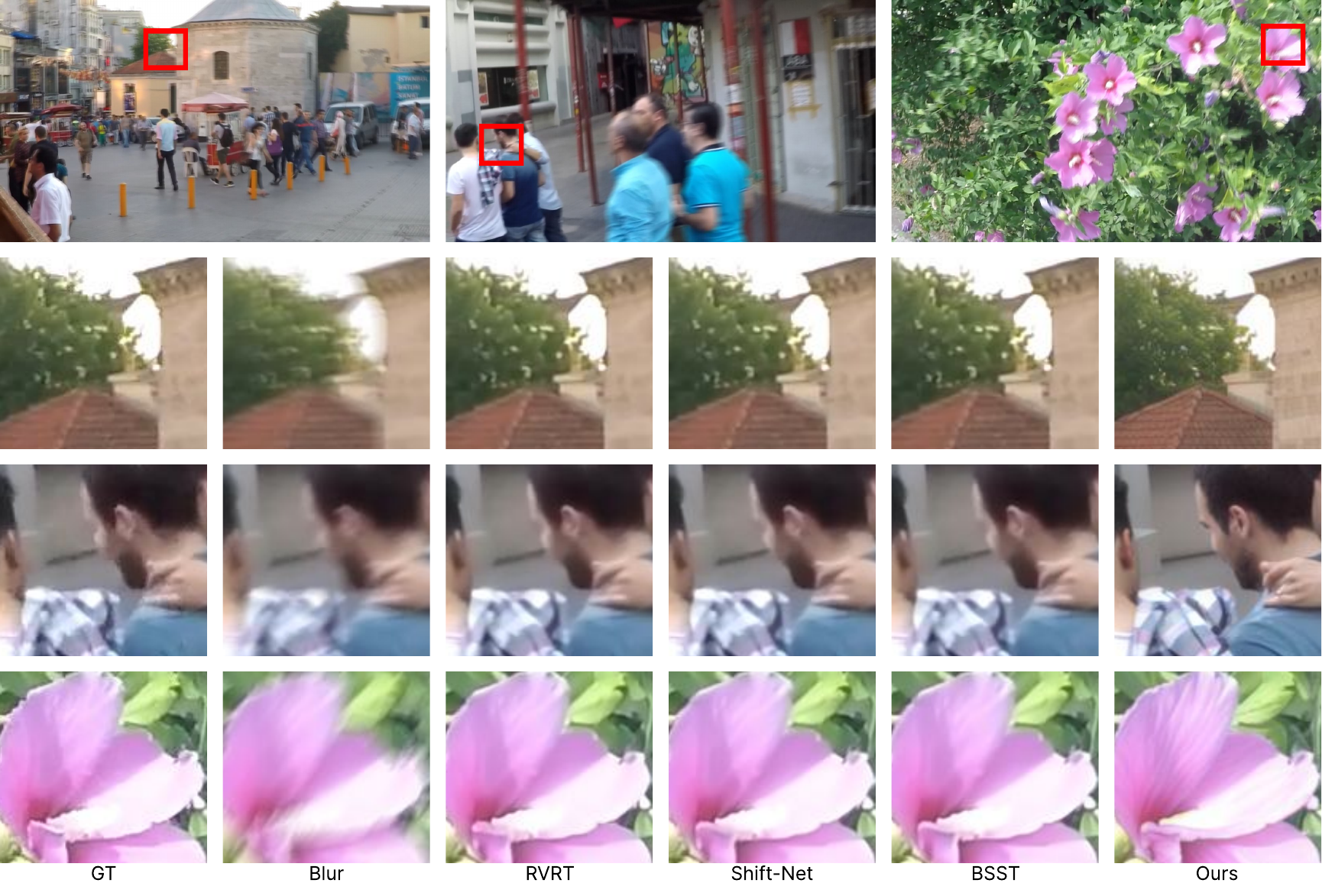}
  \caption{ Qualitative results of our method and other competitive methods on the Gopro~\cite{gopro} test set. ~\textbf{Best viewed by zooming in.}
}
  \label{fig_gopro}
\end{figure*}

\begin{table*}[ht]
  \centering
  \resizebox{\textwidth}{!}
  {%
      \begin{tabular}{c|c|ccccccccc} 
        Dataset & Method & PVDNet~\cite{pvdnet} & TSP~\cite{tsp} & STDAN~\cite{stdanet} &  DSTNet~\cite{dst} & VRT~\cite{vrt} & RVRT~\cite{rvrt} &ShiftNet~\cite{shiftnet} & BSST~\cite{bsst} & Ours  \\
        \hline

        &SSIM $\uparrow$ & 0.8705 & 0.9279 & 0.9375 & 0.9679 & 0.9724 & 0.9738 & \underline{0.9790} & \textbf{0.9792}  & 0.7808
        \\
        &PSNR $\uparrow$ & 29.20 & 31.67 & 32.62 & 34.16 & 34.81 & 34.92 & \underline{35.88}  & \textbf{35.98} & 26.80 
        \\
        \centering Gopro~\cite{gopro} &LPIPS $\downarrow$ & \underline{0.0283} & 0.0307 & 0.0289 & 0.0304 & 0.0338 & 0.0339 & 0.0347 & 0.0349 & \textbf{0.0260} 
        \\
        &NIQE $\downarrow$  & 5.602  & 5.374 & 5.377 & \underline{4.980} & 5.038 & 5.059 & 5.007 & 5.017  & \textbf{3.500}
        \\
        &MUSIQ $\uparrow$ & 36.603 & 42.766 & 42.117 & 45.379 & 45.749 & 46.283 & \underline{47.253} & 47.231 & \textbf{60.630}
        \\
        \hline
        
        &SSIM $\uparrow$ & 0.9229 & 0.9368 & 0.9374 & 0.9615 & 0.9651 & 0.9655 & \underline{ 0.9690} & ~\textbf{0.9703} & 0.7809 \\
        &PSNR $\uparrow$ & 31.87 & 32.30 & 33.05 & 33.79 & 34.27 & 34.30 & \underline{34.69} & ~\textbf{34.95}  & 26.80
        \\
        \centering DVD~\cite{dvd} &LPIPS $\downarrow$  & \underline{0.0332} & 0.0336 & 0.0344  & 0.0352 & 0.0354  & 0.0357   & 0.0345 & 0.0363 & \textbf{0.0279}
        \\
        &NIQE $\downarrow$ & 5.094 & 5.193 & 5.044 & 4.945 & 4.957 & 4.830 & \underline{4.803}  & \underline{4.803} & ~\textbf{3.513}
        \\
        &MUSIQ $\uparrow$ & 51.979 & 53.847 & 55.129 & 55.923 & 55.929 & 56.811 & 50.098 & \underline{57.063} & ~\textbf{61.135}
        \\
        \hline

      \end{tabular}%
  }
  \caption{Quantitative comparison on GoPro~\cite{gopro} and DVD~\cite{dvd} dataset. The best and the second-best results are highlight in \textbf{bold} and \underline{underline}.}
  \label{tab_quantitative}
\end{table*}

\begin{figure*}[!ht]
  \centering
  \includegraphics[width=1\linewidth]{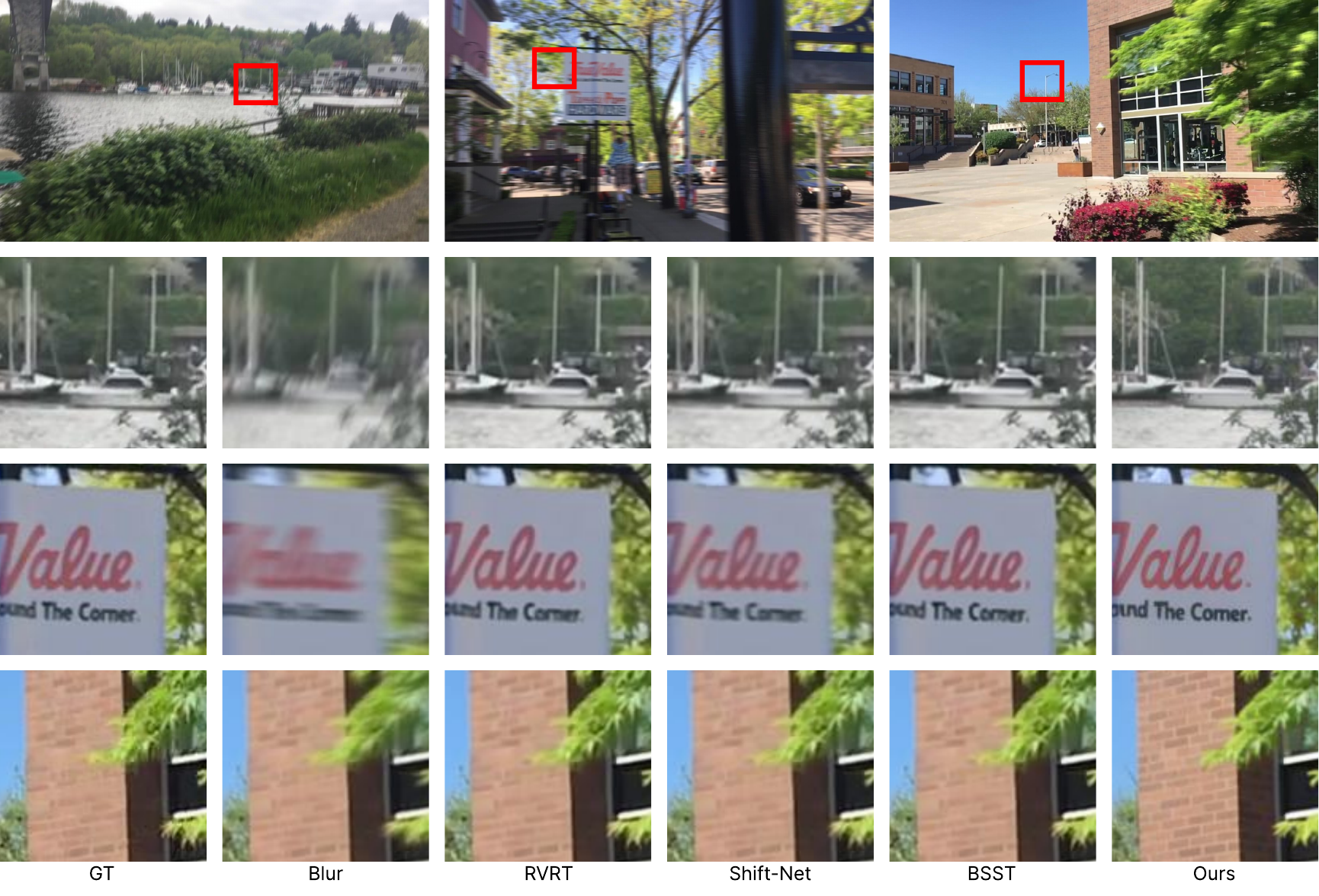}
  \caption{ Qualitative results of our method and other competitive methods on the dvd~\cite{dvd} test set. ~\textbf{Best viewed by zooming in.}
}
  \label{fig_dvd}
\end{figure*}

\subsection{Quantitative Results}

We evaluated the performance of our CPGD-Net and other sota methods on both traditional metrics (PSNR, SSIM and LPIPS~\cite{lpips}) and several no-reference image quality assessment (IQA) metrics (NIQE~\cite{niqe} and MUSIQ~\cite{musiq}). As shown in Tables~\ref{tab_quantitative}, our CPGD-Net achieves sota performance in perceptual metrics. 

We attribute the observed decrease in PSNR/SSIM values  to two key factors: First, traditional full-reference metrics like PSNR and SSIM often fail to accurately assess the quality of generative outputs. While our method produces visually superior results with more realistic textures and details, as demonstrated in Figure~\ref{fig_psnr}. Our method reconstructs significantly enhanced texture details in the woman's hair, while maintaining sharper shoulder contours without the boundary blurring artifacts observed in baseline methods in the first row. Additionally, the clothing folds of background exhibit clearer structural definition. For the second row, the foliage demonstrates markedly improved leaf clarity, and our approach generates perceptually sharper edges along the flowerbed's ridges compared to competitive. These perceptual improvements are not properly quantified by pixel-wise measurements. This phenomenon is widely recognized in recent literature~\cite{psnr1,psnr2,psnr3,psnr4}. 


Our improvements in no-reference metrics demonstrate CPGD-Net's effectiveness: a 27-30\% NIQE reduction (to 3.500/3.513) indicates better natural image statistics preservation and artifact suppression, while 7-28\% MUSIQ gains (reaching 60.630/61.135) correlate with human perception of higher quality. Combined with LPIPS scores of 0.0260/0.0279, these results show our method successfully handles video deblurring tasks while achieving superior subjective visual quality.

\subsection{Qualitative Results}



We provide a subjective comparison with previous sota models. The comparative results on the GoPro~\cite{gopro} and DVD~\cite{dvd} datasets are illustrated in Figure~\ref{fig_gopro} and Figure~\ref{fig_dvd}, respectively. The findings demonstrate that our method not only achieves effective deblurring but also enhances high-frequency details, producing visually appealing and realistic results.

Previous approaches, leveraging motion compensation algorithms based on video information and self-attention models, perform well in deblurring tasks. Similarly, our method efficiently accomplishes deblurring by incorporating coding priors. However, through comparative analysis, we observe that existing methods often lack high-frequency details, leading to outputs with reduced sharpness. In contrast, our CPC  module effectively mitigates this issue by synthesizing plausible generative details, thereby compensating for missing high-frequency components. Across diverse scenarios—including humans, clothing, nature, architecture, and text—our approach delivers sharper and more visually pleasing results.

It should be noted that the DVD~\cite{dvd} dataset stores images in JPEG format, which introduces additional encoding artifacts while reducing storage requirements. Benefiting from our hybrid high-quality raw video data and the large model's inherent real-world priors, our method can generate content beyond the constraints of ground truth, achieving superior subjective performance across various scenarios.

\subsection{Ablation Studies}
\subsubsection{The effectiveness of CPFP}

\begin{figure}[t]
  \centering
  \includegraphics[width=1\linewidth]{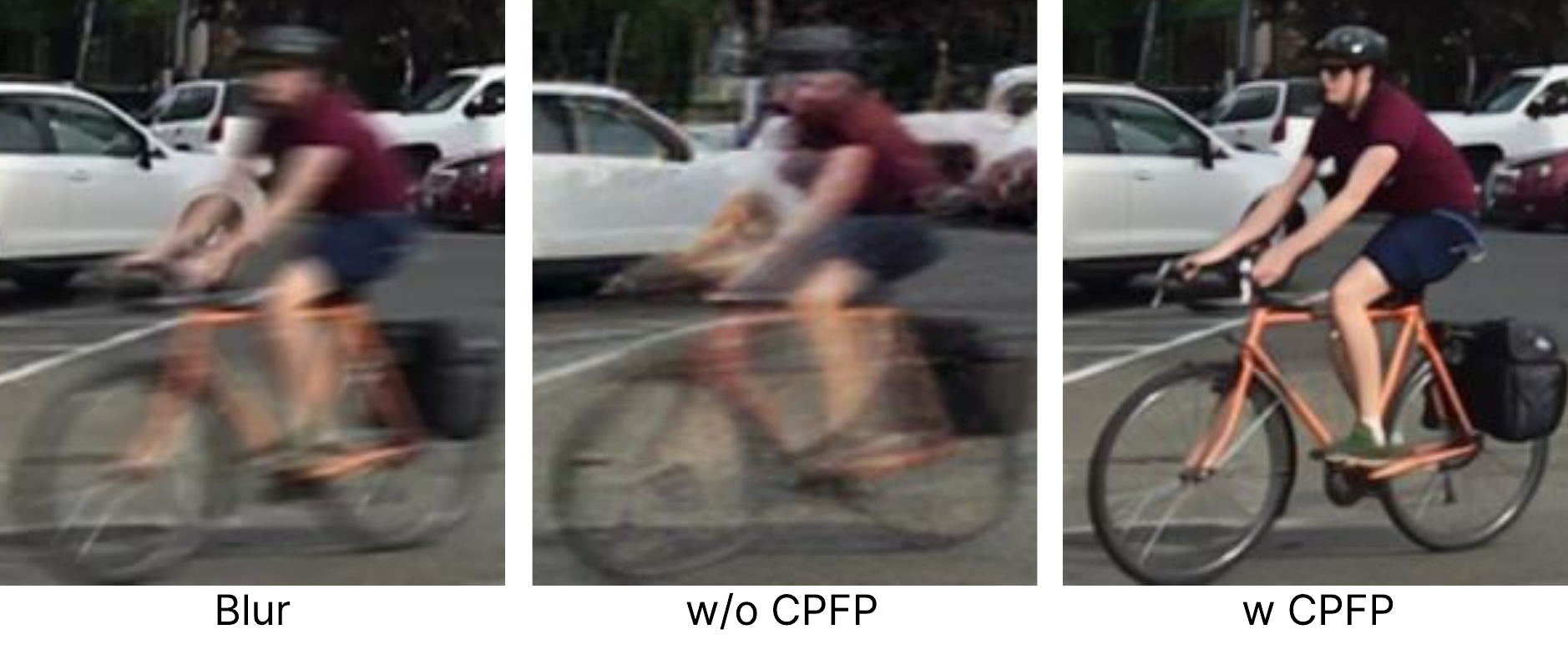}
  \caption{Ablation study of CPFP. Generative stage alone cannot reconstruct heavily blurred areas without CPFP's preliminary feature alignment.  }
  \label{fig_ab_cpfp}
\end{figure}


In this section, we validate the efficacy of our two-stage framework. First, we remove the first-stage CPFP module. As shown in Table~\ref{tab_ab_moduls}, the comprison between the first line and the last line, the quantitative results exhibit significant degradation (from 61.135 to 32.252 in MUSIQ and 3.5133 to 5.3952 in NIQE). Visual comparisons (Figure~\ref{fig_ab_cpfp}) further demonstrate that without the CPFP module, the generative model fails to restore severely blurred regions, producing outputs that remain indistinct. In contrast, the complete two-stage framework successfully accomplishes both deblurring and quality enhancement. These results collectively confirm the critical role of the CPFP module in our pipeline.

\begin{table}[t]
  \centering
  \resizebox{\linewidth}{!}
  {
      \begin{tabular}{c|cccc} 
        Modifications &w/o CPFP &w/o CPC & w IRControlNet~\cite{diffbir} & Ours\\
        \hline
        NIQE$~\downarrow$ & 5.3952 & 4.8097 & 5.0395 & 3.5133 \\
        MUSIQ$~\uparrow$ & 32.252 & 57.0717 & 56.5850 & 61.135 \\
        \hline
      \end{tabular}
    }  
  \caption{Results of ablation study. Please refer to text for details. w/o donates without.}
  \label{tab_ab_moduls}
\end{table}

\subsubsection{The effectiveness of coding prior in CPFP}

\begin{table}[t]
  \centering
  \resizebox{0.9\linewidth}{!}
  {
      \begin{tabular}{c|cc|c} 
        Modifications &NIQE$~\downarrow$ &MUSIQ$~\uparrow$ & Flops\\
        \hline
        w/o CRs &5.1102 &53.9717 \\ 
        \hline
        MVs &4.8102 &56.6301 &0\\
        RAFT~\cite{raft} Iter:4 &5.0054 &54.4335 &307G\\
        RAFT~\cite{raft} Iter:4 Init:MVs &4.8096 &57.0163 &307G\\
        RAFT~\cite{raft} Iter:20&4.8034 &57.0483 &908G\\
        \hline
      \end{tabular}
    }  
  \caption{Results of ablation study on coding priors in CPFP. w/o donates without.}
  \label{tab_ab_cpfp}
\end{table}

The coding priors consist of motion vectors (MVs) and coding residuals. First, we compare MVs with the widely-used recurrent optical flow algorithm RAFT~\cite{raft}. Additionally, since warm-start is a crucial concept in optical flow estimation, we simultaneously evaluate using MVs as initialization for RAFT. As demonstrated in Table~\ref{tab_ab_cpfp}, we compare MVs against optical flow under different configurations. The results show that optical flow performance improves significantly with increased iteration counts, with 20 iterations (908G GFLOPs for computation) yielding the best results in our ablation studies. Notably, MVs achieve competitive final performance (NIQE:4.8102, MUSIQ:56.6301) without requiring additional computational resources (0 GFLOPs). The experiments clearly demonstrate that using MVs as warm-start for RAFT leads to quality improvements - satisfactory results approaching 20-iteration performance can be obtained with just 4 iterations(NIQE:4.8096 vs 4.8034), while substantially reducing computational overhead (307 GFLOPS vs908 GFLOPS). Compared to optical flow with single iteration initialization, MVs provide superior quality. Collectively, these findings confirm that employing MVs in CPFP maintains competitive performance while significantly reducing optical flow computation.

Furthermore, we evaluate the impact of residual masks in CPFP. As shown in the first row of Table~\ref{tab_ab_cpfp}, with out coding residuals yields noticeable quality degradation from 4.81 to 5.11 in NIQE and from 56.6301 to 53.9717 in MUSIQ. The comparative results verify that residuals enhance overall performance by effectively compensating for motion estimation inaccuracies.

\subsubsection{The effectiveness of CPC}

\begin{figure}[t]
  \centering
  \includegraphics[width=1\linewidth]{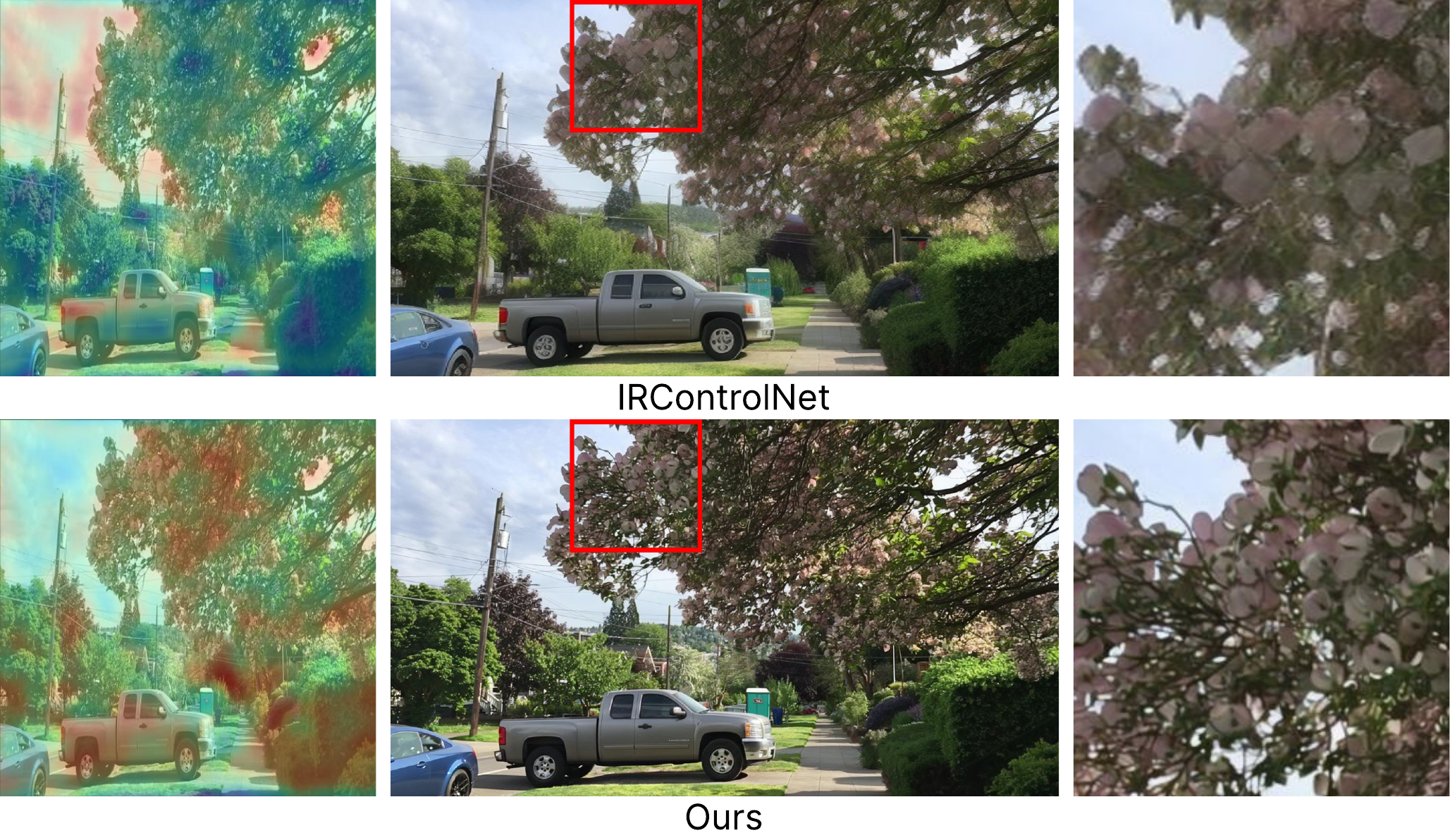}
  \caption{Visual comparison: attention heatmaps (left), outputs (middle), and zoomed details (right).}
  \label{fig_ab_heatmap}
\end{figure}

We conduct comparative experiments with IRControlNet~\cite{diffbir}, using identical first-stage outputs as input to ensure fairness since IRControlNet lacks deblurring capability. As evidenced by Table~\ref{tab_ab_moduls} (last two rows), our CPC module achieves significant performance gains, with improvement of 30.3\% in NIQE (3.5133 vs 5.0395) and 8.0\% in MUSIQ (61.135 vs 56.5850). Figure~\ref{fig_ab_heatmap} demonstrates a comparative analysis between our method and baseline method~\cite{diffbir}. The attention heatmaps (left) reveal that our method effectively focuses on critical regions—such as moving branches and vehicles—while the baseline prioritizes sky areas. The output images further substantiate our method's superior perceptual quality, zoomed-in comparisons clearly show that our approach generates sharper textures with finer details, whereas the baseline produces overly smoothed results. Both quantitative metrics and visual assessments confirm that our method outperforms the baseline in producing more natural details through targeted enhancement of motion/blur regions via encoding priors.

\subsection{Effectiveness of additional dataset}

\begin{table}[t]
  \centering
  \resizebox{0.7\linewidth}{!}
  {
      \begin{tabular}{c|cc} 
        Modifications & w/o additional dataset & Ours\\
        \hline
        NIQE$~\downarrow$ & 3.5661 & 3.5133 \\
        MUSIQ$~\uparrow$  & 55.1294 & 61.135 \\
        \hline
      \end{tabular}
    }  
  \caption{Results of ablation study. Please refer to text for details. w/o donates without.}
  \label{tab_ab_dataset}
\end{table}

\begin{figure}[t]
  \centering
  \includegraphics[width=\linewidth]{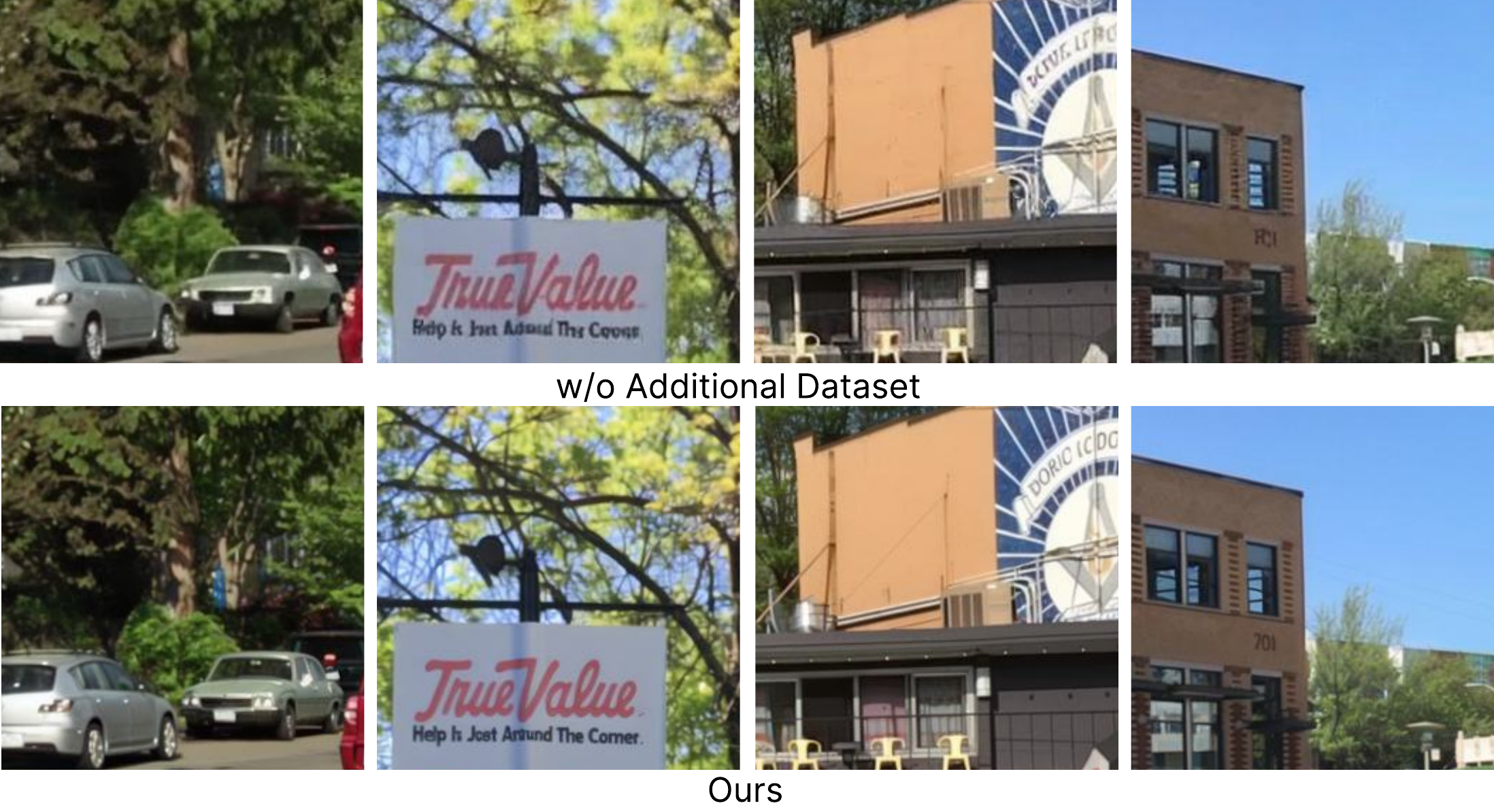}
  \caption{Visual comparison on the Effectiveness of additional dataset. ~\textbf{Best viewed by zooming in.}}
  \label{fig_ab_dataset}
\end{figure}


Without additional training data, our model tends to overfit the video deblurring dataset, causing its outputs to converge toward highly similar results due to the loss function's strong guidance toward the control images. This leads to insufficient generative detail as the optimization prioritizes direct replication of reference frames rather than synthesizing new texture information. Our comparative experiments in Table~\ref{tab_ab_dataset} and Figure~\ref{fig_ab_dataset} demonstrate that incorporating supplementary training data significantly improves output quality, with both quantitative metrics and visual results showing enhanced sharpness and detail compared to the baseline approach.

\subsection{Conclusion}

In this paper, we proposed CPGD-Net, a novel two-stage framework for video deblurring that effectively integrates coding priors (motion vectors and coding residuals) with diffusion model priors. Our approach introduces two key innovations: a Coding-Prior Feature Propagation module for efficient feature alignment and a Coding-Prior Controlled Generation module that guides a diffusion model to enhance perceptual quality. Experiments demonstrate sota performance, with 28\%/7\% MUSIQ and 30\%/27\% NIQE improvements on benchmark datasets, validating that coding priors provide valuable motion information while diffusion models can significantly enhance restoration quality when properly conditioned. This work establishes a new paradigm for combining compression-domain information with generative priors for video deblurring task. 

 \bibliographystyle{IEEEtran}
 \bibliography{sample-base}

\end{document}